\documentclass[a4paper]{article}

\usepackage{amsmath,graphicx}
\usepackage{color}

\def\x{{\mathbf x}}
\def\s{{\mathbf s}}
\def\f{{\mathbf f}}
\def\h{{\mathbf h}}
\def\L{{\cal L}}
\def\Y{{\cal Y}}
\def\R{{\cal R}}

\usepackage{amsfonts}

\usepackage{INTERSPEECH2020}

\title{Neural Zero-Inflated Quality Estimation Model\\For Automatic Speech Recognition System}
\name{Kai Fan$^*$\thanks{* indicates equal contribution.} 
\quad Bo Li$^*$ \quad Jiayi Wang$^*$ \quad Shiliang Zhang \quad Boxing Chen \quad Niyu Ge \quad Zhijie Yan}
\address{Alibaba Group Inc.}
\email{\{k.fan,shiji.lb,joanne.wjy,sly.zsl,boxing.cbx,niyu.ge,zhijie.yzj\}@alibaba-inc.com}

\begin{document}
\ninept

\maketitle
\begin{abstract}
The performances of automatic speech recognition (ASR) systems are usually evaluated by the metric word error rate (WER) when the manually transcribed data are provided, which are, however, expensively available in the real scenario. 
In addition, the empirical distribution of WER for most ASR systems usually tends to put a significant mass near zero, making it difficult to simulate with a single continuous distribution. 
In order to address the two issues of ASR quality estimation (QE), we propose a novel neural zero-inflated model to predict the WER of the ASR result without transcripts. 
We design a neural zero-inflated beta regression on top of a bidirectional transformer language model conditional on speech features (speech-BERT). 
We adopt the pre-training strategy of token level masked language modeling for speech-BERT as well, and further fine-tune with our zero-inflated layer for the mixture of discrete and continuous outputs. 
The experimental results show that our approach achieves better performance on WER prediction compared with strong baselines.
\end{abstract}

\section{Introduction}
\label{sec:intro}

Automatic Speech Recognition (ASR) has made remarkable improvement since the advances of deep learning \cite{hinton2012deep} and powerful computational resources \cite{Chen2018}. 
However, current ASR systems are still not perfect because of the constraint of objective physical conditions, such as the variability of different microphones or background noises.
Thus, quality evaluation (QE) is a practical desideratum for developing and deploying speech language technology, enabling users and researchers to judge the overall performance of the output, detect bad cases, refine algorithms, and choose among competing systems in a specific target environment. 
This paper focuses on such a situation where the golden references are not available for estimating a reliable word error rate (WER). 

\begin{figure}[t]
\includegraphics[width=\columnwidth]{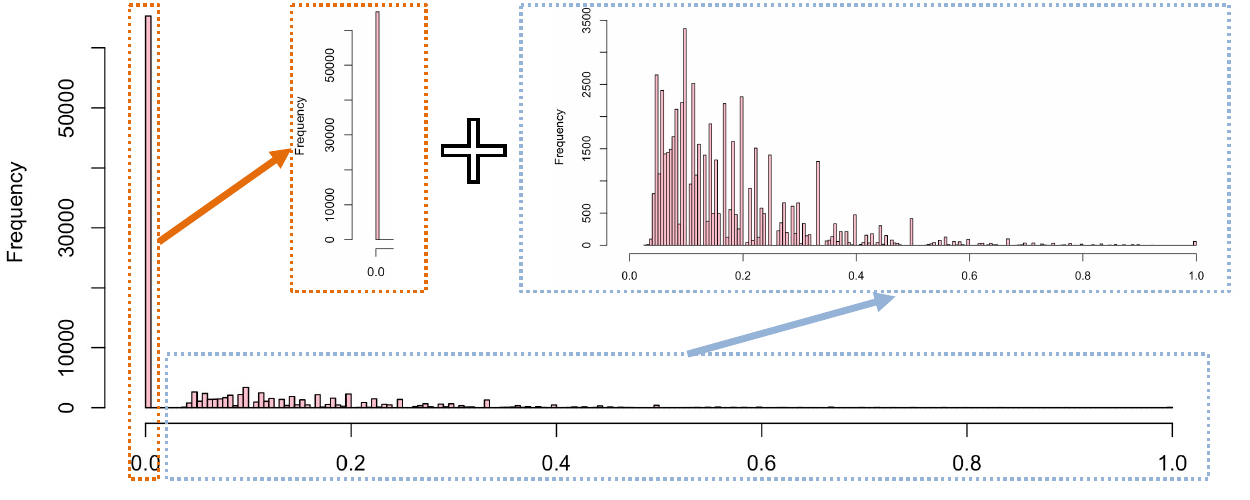}
\caption{The empirical distribution of WER for our ASR system. The majority of WER values are concentrated at 0, and the rest approximately follows a Gaussian or Beta distribution.}
\label{fig:spike}
\end{figure}

In the research direction of ASR QE without transcripts, a two-stage framework including feature extraction and WER prediction, has been a long-standing criteria. 
Classical pioneering works mainly rely on hand-crafted features \cite{le2018advanced} and utilize them to build a linear regression based algorithms, including aggregation method with extremely randomized trees \cite{sperber-etal-2016-lightly}, SVM based TranscRater \cite{jalalvand2016transcrater}, e-WER \cite{ali2018word}, and error type classification \cite{ogawa2012error,ogawa2015asr}. 
In this work, instead of heavily using manual labors, we propose to derive the feature representations from a pre-trained conditional bidirectional language model -- speech-BERT, which aims to predict the relationships between the raw fbank features and utterances by analyzing them holistically. 
The training data required for speech-BERT is exactly the same as the one for conventional ASR, without any additional human annotations. 
Subsequently, during the WER prediction stage, we analyze its empirical distribution for most ASR systems (in Fig.~\ref{fig:spike}), and find that WER values are prone to distribute near 0. 
Therefore, during the fine-tuning procedure, we introduce a neural zero-inflated regression layer on top of the speech-BERT, fitting the target distribution more appropriately.

In summary, this paper makes the following main contributions. 
\textbf{i)} We propose a bidirectional language model conditioned on speech features. The aim of this model is to improve the feature representations for ASR downstream tasks. 
A \textbf{bonus} experiment shows that tying the parameter of speech-BERT and speech-Transformer can even achieve comparable performance for both task with one single training. 
\textbf{ii)} We introduce a neural zero-inflated Beta regression layer, particularly fitting the empirical distribution of WER. 
For the gradient back-propagation of neural Beta regression layer, we design an efficient pre-computation method. 
\textbf{iii)} Our experimental results demonstrate our ASR quality estimation model can outperform many classical quality estimation models. 

\section{Related Works}
\label{sec:related}

Transformer \cite{vaswani2017attention,devlin2019bert,lample2019cross} has been extensively explored in natural language processing, and become popular in speech \cite{dong2018speech,zhang-etal-2019-lattice}. 
Our motivation comes from the success of BERT \cite{devlin2019bert}, demonstrating the importance of bidirectional pre-training for language representations and reduced the need for many heavily-engineered task specific architectures. 
We will also adopt the loss function of masked language model $p(\x_{mask}|\x_{unmask})$ as our training criteria, where $\x$ represents all tokens/utterances in one sentence. 
However, the major difference is that speech-BERT is conceptually a \textit{conditional} language model \cite{ghazvininejad2019constant}, in order to capture the subtle correlations between speech features and utterances as well as the syntactic information.

In order to build the conditional masked language model $p(\x_{mask}|\x_{unmask}, \s)$, where $\s$ is the speech features corresponding to $\x$, we have to discard the single transformer encoder architecture of BERT since it's difficult to consume two sequences of different modalities, though it's doable with XLM \cite{lample2019cross}. 
We propose to modify the speech-Transformer \cite{dong2018speech} by changing its auto-regressive decoder to a parallel memory encoder, resulting in an ``encoder-memory encoder" architecture, where the speech and text domains can be separately controlled by two different encoders. 
The memory means the outputs of the speech encoder by consuming the spectrogram inputs. 

When the feature representations are ready, the quality estimation task is typically reduced to either a regression or a classification problem, depending on the type of predicated values. 
In machine translation (MT) quality estimation \cite{bojar2016ten}, translation error rate (TER), a real-valued metric similar to WER, or tag OK/BAD, is the target of the model, and will be predicted at sentence or token level. 
However, a standard regression model is probably not suitable for fitting the WER of ASR systems. 
As mentioned before, we observe that the distribution of WER is empirically more zero-concentrated than TER's, making a straightforward linear regression easily biased. 
 
We propose a neural zero-inflated regression layer, motivated by the statistical inflated distribution \cite{ospina2010inflated,rovckova2018spike}, being capable of modeling the mixed continuous-discrete distributions. 
A random variable $y$ generated by such a distribution is typically represented by a weighted mixture of two distributions.
\begin{align}\label{eq:zero-inflated}
\lambda \cdot \left(\sum_{y_i \in \Y} p_{y_i} \mathbb{I}_{y=y_i}\right) + (1-\lambda) \cdot p_{\R / \Y}(y)
\end{align}
where $\Y$ is a finite set, and $\mathbb{I}_{y=y_i}$ is an indicator function whose value equals to 1 if $y$ equals to $y_i$. 
In our case where $y$ denotes WER, we have $\Y=\{0\}$ and $p_0=1$. 
In particular, for ASR-QE we recommend using a Beta distribution for $p(y)$. 
Then $\lambda$ simply represents the probability of whether $y$ takes the value 0 or not, resulting a mixture of Bernoulli and Beta distribution. 
Additionally, we use one classification neural network to simulate the Bernoulli variable and a regression neural network to simulate the continuous variable, resulting a differentiable deep neural architecture that can be fine-tuned together with the parameters of speech-BERT. 
In this way, we can cast the ASR-QE problem into a hierarchical multi-task learning paradigm, where the first step is to decide whether the ASR output is perfect or not, and the second step is only to regress the WER value for the imperfect one.  

\begin{figure}[t]
\includegraphics[width=\columnwidth]{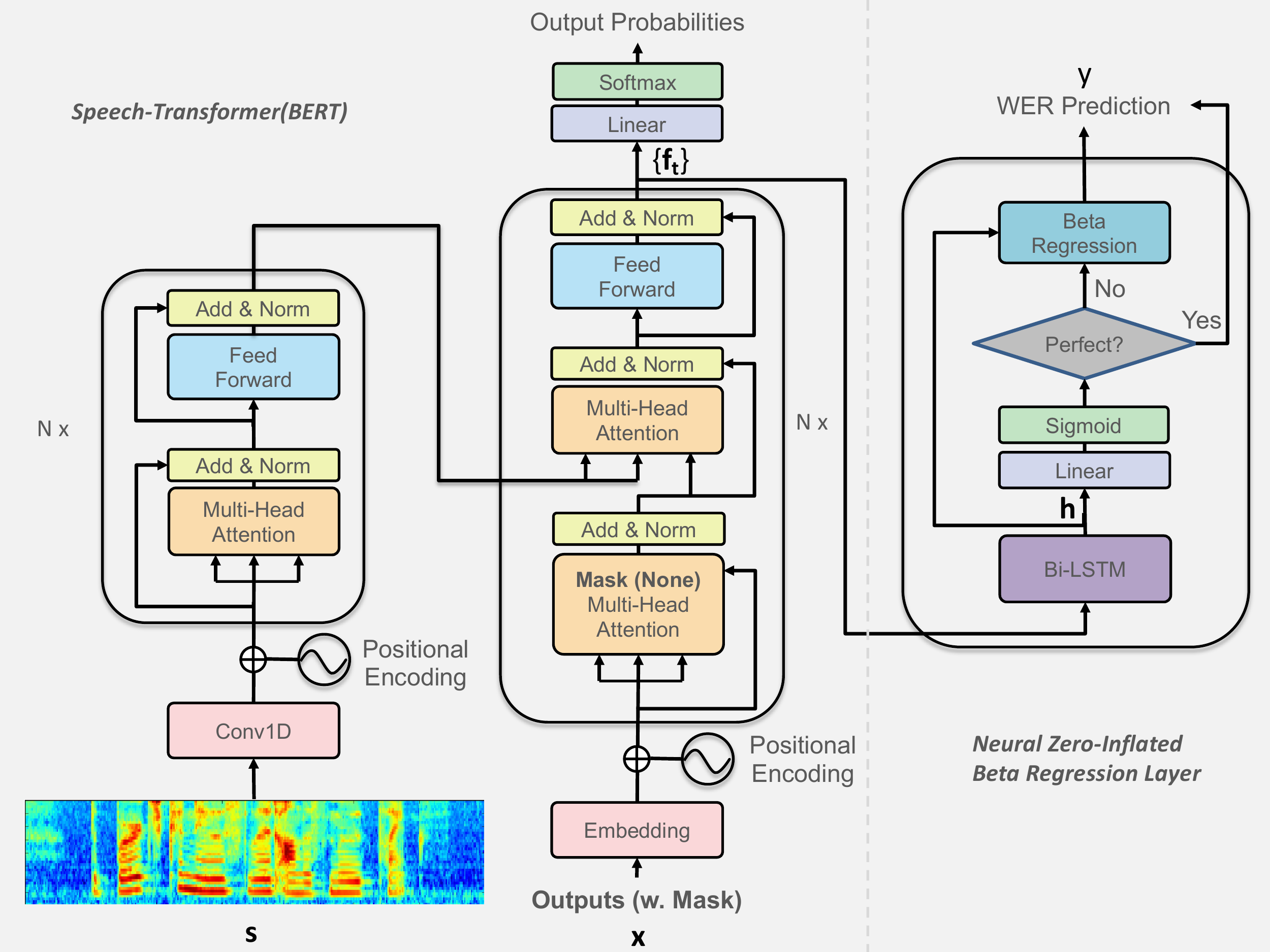}
\caption{Model architecture of speech-BERT with zero-inflated Beta regression layer. The text in brackets denotes description of the memory encoder of speech-BERT, otherwise it represents the decoder of speech-Transformer.}
\label{fig:model}
\end{figure}

\section{Methodology}
\label{sec:method}

\subsection{Speech-BERT}

The backbone structure of speech-BERT originates from adapting the decoder in the speech-Transformer \cite{dong2018speech} to a memory encoder (in Fig.~\ref{fig:model}). 
To achieve this goal, we need two simple modifications. 
First, we randomly change 15\% utterances in the transcription at each training step. 
We introduce a new special token ``[mask]" analogous to standard BERT and substitute it for the tokens required masking. 
Notice that in practice 15\% of utterances that required prediction during pre-training include 12\% masked, 1.5\% substituted and 1.5\% unchanged. 

Secondly, we also remove the future mask matrix in the self-attention of the decoder, which can be concisely written as a unified formulation $\text{softmax}\left(\frac{Q_{\x}K_{\x}^\top}{\sqrt{d/h}} + \mathbb{I}_{\text{ST}}M\right)V_{\x}$,
%
%
where the indicator $\mathbb{I}_{\text{ST}}$ equals to 1 if the model architecture is speech-Transformer. 
The $K_{\x}, Q_{\x}, V_{\x}$ are the output keys, queries, and values from the previous layer of the decoder or memory encoder. 
$M$ is a triangular matrix where $M_{ij}=-\infty$ if $i < j$. 
In the case of the decoder, the self-attention represents the information on all positions in the decoder up to and including that position. 
In the case of the memory encoder, it represents the information on all positions excepted the masked positions. 
Other details are similar to the standard transformer in \cite{vaswani2017attention}. 
The advantage of using the unified formulation is that it allows us to straightforwardly implement a multi-task learning task in a weights-tying architecture via altering the mask matrix, resulting in the summation of two losses $\L(\x,\s;\theta)=\L_{\text{SB}}(\x,\s;\theta) + \lambda_{\text{ST}}\L_{\text{ST}}(\x,\s;\theta)$, i.e.,
\begin{align}
\L(\x,\s;\theta) =& - \log p(\x_{mask}|\x_{unmask},\s;\theta) \nonumber\\
                  &- \lambda_{\text{ST}}\sum_{t} \log p(\x_t|\x_{<t}, \s;\theta) 
\end{align}
where model parameter $\theta$ may be shared cross speech-BERT and speech transformer. 
The extra ASR loss also differentiates our model from the standard BERT whose additional loss is designed for next sentence prediction (NSP) task. 
In multi-task learning, we usually set $\lambda_{\text{ST}} = 0.15$ to keep two different losses at a consistent scale.

\subsection{Neural Zero-Inflated Regression Layer}

The speech-BERT is able to unambiguously output a sequence of feature representations $\{\f_t\}_{t=1}^T$ corresponding to every single utterance in the transcription. 
Theoretically, we can use a single feature representation of an arbitrarily selected token for many downstream tasks like ``[CLS]" in standard BERT, since the self-attention mechanism has successfully integrated all syntactic information into every feature. 
Intuitively, it is reasonable to use another feature fusion layer to encode the sequence of features together. 
Thus, we use one Bi-LSTM \cite{graves2013hybrid} layer to re-encode the features and output a single final encode state as the feature $\h$ for the quality estimation task.
 
Referring to Eq.~(\ref{eq:zero-inflated}), we can first define a binary classifier to indicate that whether the ASR result is flawless or not, i.e., following the Bernoulli distribution $Bern(\lambda)$.
\begin{align}\label{eq:cls}
p(\text{WER}_{\x}=0) = \lambda = \texttt{sigmoid}(\mathbf{w}_{\lambda}^\top \h + b_{\lambda})
\end{align}

For the subsequent regression model, it is not necessary to predict the case of zero WER due to the existence of above classifier. 
This fact naturally advocates the choice of Beta regression because the Beta distribution has no definition on zero. 
For statistical distributions, the most importanct statistics are usually the first two moments, i.e., mean and variance. 
In our proposal, we mainly model the mean $\mu$ which is the actual target in our final prediction, and derive the variance of Beta distribution.
\begin{align}\label{eq:reg}
\mu = \texttt{sigmoid}(\mathbf{w}_{\mu}^\top \h + b_{\mu}), \quad \sigma = \frac{\mu(1-\mu)}{1+\phi}
\end{align}
where $\phi$ is a hyper-parameter that can be interpreted as a precision parameter, which can be estimated from the training data. 
The parameterized density distribution function is expressed as follows.
\begin{align}\label{eq:beta}
p(y) = \frac{\Gamma(\phi)y^{\mu\phi-1}(1-y)^{(1-\mu)\phi-1}}{\Gamma(\mu\phi)\Gamma((1-\mu)\phi)}, \quad 0<y<1
\end{align}
Combining Eq.~(\ref{eq:cls},\ref{eq:reg},\ref{eq:beta}), the training objective with the neural zero-inflated beta regression layer is to maximize the log-likelihood of the proposed distribution of WER.
\begin{align}
\max \log p(\text{WER}_{\x}=0) + \mathbb{I}_{\text{WER}_{\x}>0} \cdot \log p(y)
\end{align}
This is a hierarchical loss for two consecutive sub-problems but can be simultaneously optimized, where the first classification loss requires the whole fine-tune dataset, while the second term is only fed with the data of inaccurate transcriptions. 
With this loss function, we use the expected prediction during inference, i.e., $p(\text{WER}>0) \cdot y_{\text{pred}}$.

\textbf{Discussion} If $\Y$ has $K > 1$ elements, Eq.~(\ref{eq:cls}) can be easily generalized to support a categorical distribution with $K+1$ classes via \texttt{softmax}. 
With a more succinct parameterization than the Eq.~(\ref{eq:zero-inflated}), the output $K + 1$ classes actually represent the probabilities, $1-\lambda$ and $\lambda p_{y_i}, i=1,...,K$, where $\lambda$ is absorbed into $p_{y_i}$ as a single parameter. 

\subsection{Gradient Pre-Computation}
A crucial issue for Beta regression layer is the involved gradient computation of $\log p(y)$ is not straightforward, since the direct auto-differentiation with respect to the training objective is obliged to calculate the gradient of a compound Gamma function $\Gamma(g_{\mu}(\x,\s;\theta_{\mu}))$, where $g_{\mu}$ simply denotes the computational graph or function with the input $\x,\s$ and the output $\mu$. 
Instead, we utilize a gradient pre-computation trick so that the back-propagation becomes less cumbersome. 
To do so, we propose an equivalent objective $\tilde{\L_p}$ as to log-likelihood $\log p(y)$.
\begin{align}
\tilde{\L_p} &= \phi \cdot g_{\mu} \cdot \texttt{stop\_gradient}(y^* - \mu^*)
\end{align}
where $y^*= \log y / \log(1-y), \mu^* = \psi(\phi\cdot g_{\mu}) - \psi(\phi\cdot (1 - g_{\mu}))$ and $\psi(x) = \frac{\mathrm{d}}{\mathrm{d}x}\log\Gamma(x)=\frac{\Gamma'(x)}{\Gamma(x)}$ is digamma function (e.g., $\texttt{tf.digamma}$ in TensorFlow). 
The equivalence is essentially in the sense of gradient operator. 
In other words, the stochastic gradient optimization will still remain the same because we can derive the following identical relation by algebra calculations. 
\begin{align}
\nabla_{\theta_{\mu}} \tilde{\L_p} = \nabla_{\theta_{\mu}} \log p(y)
\end{align}
In the new objective, we have successfully circumvented the direct gradient back-propagation with respect to $\Gamma(g_{\mu}(\cdot))$, since the complicated term $y^* - \mu^*$ merely involves forward digamma function computation with a stop gradient operation, while the term $g_{\mu}$ can readily and efficiently contribute to the back-propagation because it simply consists of the common operations in deep neural networks.

\section{Experiments}
\label{sec:exp}

For validating the effectiveness of our approach, the quality estimation model of ASR was evaluated by three metrics,  normalized discounted cumulative gain (NDCG), Pearson correlation and Mean Absolute Error (MAE). 

\subsection{Baseline Benchmarking}
First, we conduct an experiment to compare with the popular open-source ASR-QE tool Transcrater\cite{jalalvand2016transcrater}.
We use the exactly same data from 3rd CHIME challenge \cite{barker2015third} and the same evaluation metrics described in the original paper. 
Note that we adopt the convention of Transcrater by multiplying 100\% for MAE.  
Language Model (LM) and Part-of-speech (POS) features in Transcrater are trained on external large scale data  \cite{mikolov2010recurrent,stolcke2000sri,schmid2013probabilistic}. 
However, we didn't use any external data in our algorithm, and set a relatively small model size (3 layers / 256 hidden units) to prevent overfitting. 
8738 noisy utterances from a total 4 speakers in the real data are used as pre-training, while 1640 utterances are used for fine-tuning QE model. 
Table~\ref{tab:benchmark} shows the detailed comparison between our algorithms and Transcrater.

\begin{table}[h]
\caption{MAE results benchmarking}
\setlength{\tabcolsep}{2pt}
\centering
\begin{tabular}{c|c|c|c}
\hline
Method & Features & MAE$\downarrow$ & NDCG$\uparrow$ \\
\hline
Transcrater & Baseline & 28.7 & 73.6 \\
	             & SIG & 27.3 & 73.5 \\
 & LEX/LM/POS (external data) & 22.2 & 80.4 \\
 & SIG/LEX/LM/POS (external data) & 23.5 & 79.4 \\
\hline
Ours & no pre-training & 10.2 & 76.2 \\
	 & with pre-training & \textbf{9.87} & \textbf{84.0}\\
\hline
\end{tabular}
\label{tab:benchmark}
\end{table}

\subsection{Large Scale Dataset}
The second experiment is to test the model performance when the dataset is largely scaled. 
A large Mandarin speech recognition corpus containing 20,000 hours of training data with about 20 million sentences is used for speech-BERT pre-training. 
We evaluate the performance of pre-training via the prediction accuracy on masked tokens. 
The dataset of quality estimation includes 240 hours, which do not appear in the pre-training dataset. 
The speech recognition system that we want to evaluate the quality on is an in-house ASR engine based on Kaldi\cite{povey2011kaldi}. 
The WER computed from the ASR results and the ground truth transcripts is the target our model will predict.
We have two test sets of the quality estimation model for in-domain and out-of-domain, each of which contains 3000 sentences. 
The acoustic features used in our model are 80-dimensional log-mel filter-bank (fbank) energies computed on 25 ms window with 10 ms shift. 
For computational efficiency for the speech encoder, we stack the consecutive frames within a context window of 4 to produce the 320-dimensional features. 
The speech-BERT is trained on 8 Tesla V-100 GPUs for about 10 days until convergence. 
The quality estimation model is fine-tuned on 4 Tesla V-100 GPUs for several hours. 

\subsection{Pre-Training Results}

\begin{figure}[t]
\includegraphics[width=\columnwidth]{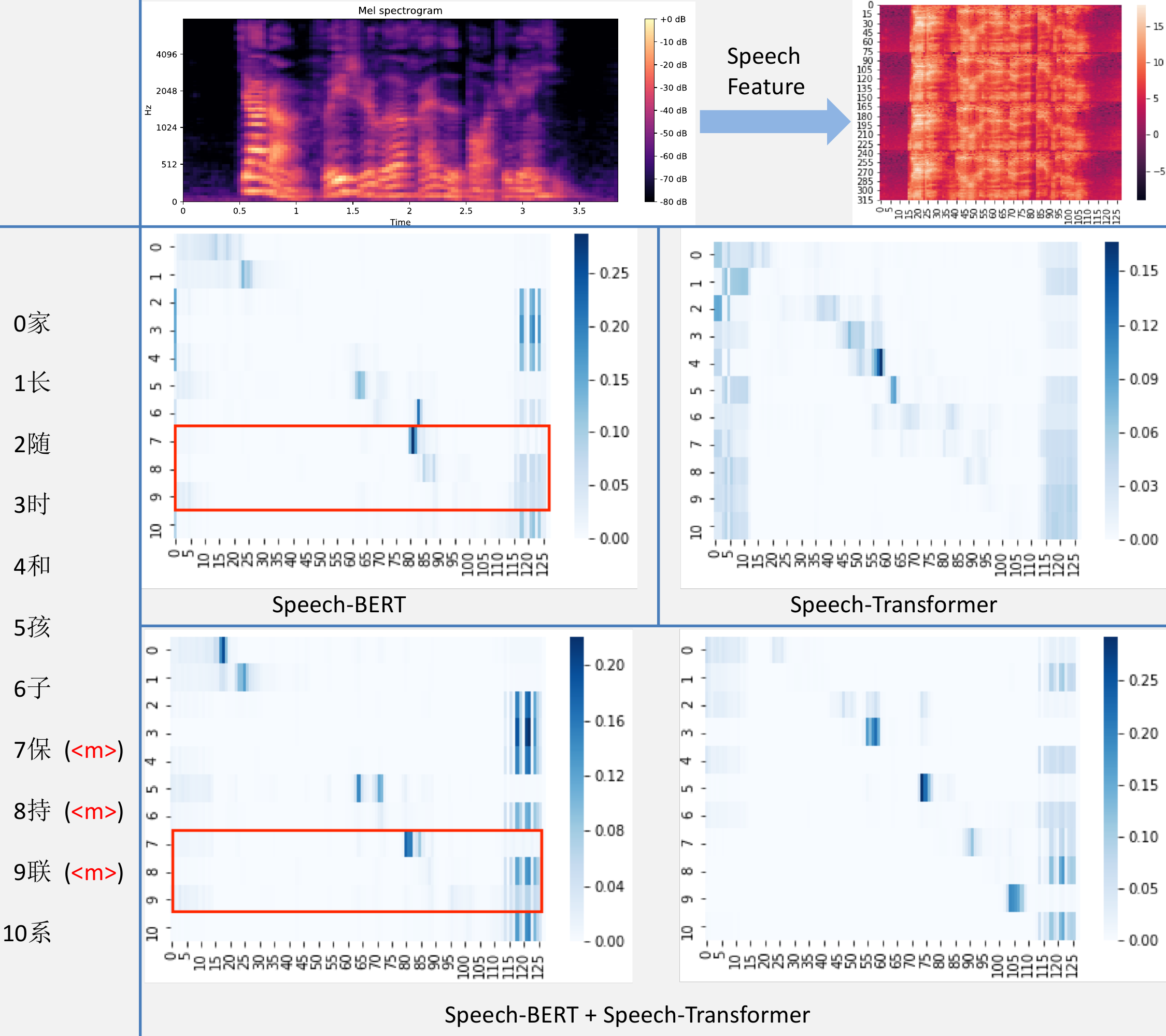}
\caption{An example of attention visualization for different pre-training losses.}
\label{fig:attention}
\end{figure}

We train the speech-BERT model with three different loss functions and the results are summarized in Table~\ref{tab:pre-training}. 
We observe that the jointly trained model can achieve comparable performance to that of two separately trained tasks. 
We also visualize the attention between the speech encoder and the memory encoder in Fig.~\ref{fig:attention}. 
The attention weights are averaged over 8 heads, and the overall patterns of joint training and separate training are relatively similar. 
We prefer to adopt the simultaneously pre-trained model as our downstream quality estimation task, since we hypothesize that the more supervisions in multi-task learning may incorporate more syntactic information in the hidden representations. 

\begin{table}[h]
\caption{The performance of pre-training}
\centering
\begin{tabular}{c|c|c|c}
\hline
Loss & $\L_{\text{SB}}$ & $\L_{\text{ST}}$ & $\L_{\text{SB}} + \L_{\text{ST}}$ \\
\hline
Masked Token Predict Acc & 95.39\% & NA & 94.81\% \\
WER (beam=5) & NA & 9.23 & 10.14 \\
\hline
\end{tabular}
\label{tab:pre-training}
\end{table}

\begin{figure}[t]
\includegraphics[width=\columnwidth]{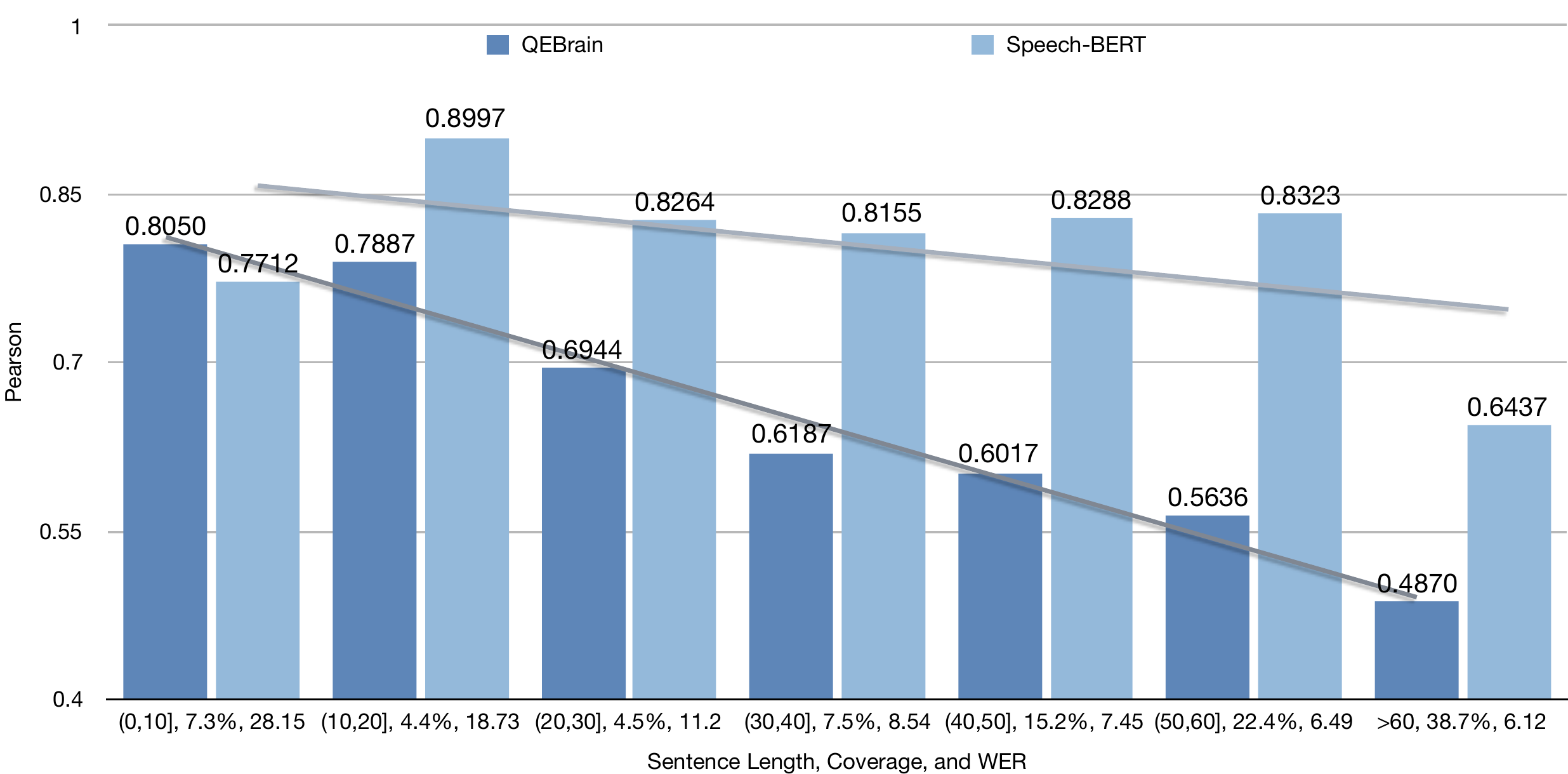}
\caption{Pearson coefficient for different sentence lengths. The overall WER is 8.98 for our in-house Kaldi ASR engine.}
\label{fig:comparison}
\end{figure}

\begin{table}[t]
\caption{The comparison of different last layers for quality estimation model. Each setting is run 4 times.}
\centering
\begin{tabular}{c|c}
\hline
Last Layer & Pearson$\uparrow$ \\
\hline
Zero-Inflated Beta Regression & 0.5786 $\pm$ 0.0041 \\
Linear Regression & 0.5486 $\pm$ 0.0086 \\
Zero-Inflated Linear Regression & 0.5738 $\pm$ 0.0019 \\
Logistic Regression & 0.5501 $\pm$ 0.0006 \\
Zero-Inflated Logistic Regression & 0.5726 $\pm$ 0.0061 \\
\hline
\end{tabular}
\label{tab:zero-beta}
\end{table}

\begin{table}[t]
\vspace{-2mm}
\caption{The comparison of different quality estimation models.}
\centering
\begin{tabular}{c|c|c|c}
\hline
Method & MAE$\downarrow$ & Pearson$\uparrow$ & F1$\uparrow$ \\
\hline
QEBrain     & 7.30 & 0.7829 & 0.4956 \\
Ours & \textbf{5.60} & \textbf{0.8187} & \textbf{0.5372} \\
\hline
\end{tabular}
\label{tab:qe}
\end{table}

\subsection{Fine-Tuning Results}

For the quality estimation model, we first explore the advantage of zero-inflated model and Beta regression by varying $p(y)$, the prior distribution of WER. 
The performances of different settings in the last layer are evaluated on the out-of-domain test set and are shown in Table~\ref{tab:zero-beta}. 
Notice that \textbf{i)} we cannot simply apply Beta regression to the last layer without zero-inflation, since the zero WER will violate the support of Beta distribution. 
\textbf{ii)} The linear regression does not necessarily mean a pure Gaussian distribution, since the output still has to be confined in the interval $[0,1]$. 
Thus, the mean of the Gaussian distribution cannot be arbitrarily large but should be applied a sigmoid function in advance. 
\textbf{iii)} As the description in ii), the logistic regression is merely different from linear regression in the loss functions (cross-entropy v.s. mean squared loss). 
\textbf{iv)} The precision parameter of Beta regression is a hyper-parameter estimated from the training data satisfying $\text{WER}>0$. 
We employ the maximum likelihood estimation with another common parameterization $\phi=a+b$, where $\frac{\Gamma(a+b)y^{a-1}(1-y)^{b-1}}{\Gamma(a)\Gamma(b)}$.

Another experiment we conduct on our in-domain test dataset is to compare our ASR-QE model as an integrated pipeline with the state-of-the-art quality estimation model QEBrain\cite{fan2019bilingual} in WMT2018 \cite{specia2018findings}. 
Notice that for fair comparison, we have to modify the text encoder of QEBrain to be exactly same speech encoder as ours and the last layer to a zero-inflated regression one.
In addition to the Pearson and MAE, we also introduce F1 measure as the harmonic mean of precision and recall, which is typically used in binary (0/1) classification. 
We label the recognition results ``acceptable" (i.e., 1) when the predicted WER $<=0.14$. 
The overall results illustrated in Table~\ref{tab:qe} shows that speech-BERT outperforms QEBrain on all three metrics.
Furthermore, we have a detailed analysis on the Pearson with respect to different sentence lengths in Fig.~\ref{fig:comparison} where we use two linear regression lines to fit the decreasing performance as the sentence length increases. 
QEBrain demonstrates better performance when the sentence is shorter, but speech-BERT has a stable performance across all length ranges. 
This finding makes sense because the longer sentences are likely to have lower or even near-zero WER, which can be better dealt with by zero-inflated Beta regression layer.

\section{Conclusions}
\label{sec:conl}

In this study, we first propose a deep architecture speech-BERT, which is seamlessly connected to the speech-Transformer. 
The key purpose is to pre-train the model on large scale ASR dataset, so that the last layer of the whole architecture can be directly fed as downstream features without any manual labor. 
Meanwhile, we designed a neural zero-inflated Beta regression layer, which closely models the empirical distribution of WER. 
The main intuition is to regress a variable defined as a mixture of the discrete and continuous distributions. 
With the elaborated gradient pre-computation method, the loss function can still be efficiently optimized. 
In addition, if following the recent work \cite{Swarup2019}, we can probably improve the confidence scores at the token level, which will be served as future work.


\newpage
\bibliographystyle{IEEEtran}

\bibliography{mybib}


\end{document}